\documentclass{article}

\usepackage{arxiv}

\usepackage[utf8]{inputenc} 
\usepackage[T1]{fontenc}    
\usepackage{hyperref}       
\usepackage{url}            
\usepackage{booktabs}       
\usepackage{amsfonts}       
\usepackage{nicefrac}       
\usepackage{microtype}      
\usepackage{lipsum}		
\usepackage{graphicx}
\usepackage{natbib}
\usepackage{doi}

\usepackage{soul}
\usepackage{url}
 \usepackage{wrapfig, subcaption, setspace, booktabs}
\usepackage{multirow}
\usepackage{enumitem}
\setitemize{noitemsep,topsep=0pt,parsep=0pt,partopsep=0pt}
\title{IMG-NILM: A Deep learning NILM approach using energy heatmaps}

\author{ {\hspace{1mm}Jonah Edmonds} \\
	Department of Engineering Mathematics\\
	University of Bristol, UK\\
	\texttt{je17667@bristol.ac.uk} \\
	\And
	{\hspace{1mm}Zahraa S. Abdallah} \\
	Department of Engineering Mathematics\\
	University of Bristol, UK\\
	\texttt{zahraa.abdallah@bristol.ac.uk} \\
}

\date{}

\renewcommand{\shorttitle}{IMG-NILM:A Deep learning NILM approach using energy heatmaps}

\hypersetup{
pdftitle={IMG-NILM: A Deep learning NILM approach using energy heatmaps},
pdfsubject={},
pdfauthor={Jonah Edmonds, Zahraa S. Abdallah},
pdfkeywords={First keyword, Second keyword, More},
}

\begin{document}
\maketitle

\begin{abstract}
Energy disaggregation estimates appliance-by-appliance electricity consumption from a single meter that measures the whole home's electricity demand. Compared with intrusive load monitoring, NILM (Non-intrusive load monitoring) is low cost, easy to deploy, and flexible. In this paper, we propose a new method, coined IMG-NILM, that utilises convolutional neural networks (CNN) to disaggregate electricity data represented as images. Instead of the traditional approach of dealing with electricity data as time series, IMG-NILM transforms time series into heatmaps with higher electricity readings portrayed as `hotter' colours. The image representation is then used in CNN to detect the signature of an appliance from aggregated data. IMG-NILM is robust and flexible with consistent performance on various types of appliances; including single and multiple states. It attains a test accuracy of up to 93\% on the UK-Dale dataset within a single house, where a substantial number of appliances are present. In more challenging settings where electricity data is collected from different houses, IMG-NILM attains also a very good average accuracy of 85\%.
\end{abstract}


\section{Introduction}
With the urgency to reduce the negative impact of fossil fuels, much of the focus has been directed toward  energy production, with much time and money being invested in renewable energy sources such as wind, solar and water-based generation. However, less attention has been directed towards later stages of the electricity industry, where control of usage and recycling of waste can have a major impact on the electrical requirement of the globe. Intelligent use of electricity will be the most important factor in solving the world's energy problems, according to the International Electrotechnical Commission (IEC). Hence, having a more thorough understanding of energy consumption can have a positive impact on how we address the climate crisis. The ability to disaggregate the net energy into individual appliances will enhance consumption understanding which will benefit energy market stakeholders from consumers to suppliers. Applications for disaggregation include outlier detection to spot faulty or expensive appliances, simplifying costs and bills, and spotlighting temporal trends.

An intrusive approach for energy disaggregation requires each appliance to be fitted with a meter that records the power output of that individual device before relaying that information back to the user. Whilst this method is accurate, it has many concerns in terms of privacy and cost. Alternatively, a non-intrusive Load Monitoring (NILM) takes the aggregated power consumption of the building and uses this information to estimate the individual appliance data. This approach is more practical, cheaper and ethically less controversial than the first one, therefore, being a much more viable and attractive prospect to consumers concerned about their privacy. However, NILM is currently not as accurate and therefore requires research and development to become as reliable as the installation of meters within houses. 

Deep neural networks (DNNs) have become popular in many domains such as image recognition, speech translation and automated cancer cell detection \cite{samek2021explaining}. DNNs are also applied to NILM, yet still encounter many challenges in terms of accuracy and robustness, especially when applied across different types of appliances. This paper, therefore, utilises DNNs for NILM to address some of these challenges. The contribution of this paper is summarised as follows:   
\begin{itemize}
    \item We introduce a new image representation for electricity data. A typical approach deals with electricity data as a time series, nevertheless, patterns in time series data are not intuitively detected  neither visually nor analytically. Alternatively, transforming time series data into the form of a heatmap will preserve the time dependency whilst revealing patterns related to user behaviour, routine, weather and other time-dependent factors.   
    \item IMG-NILM showed very good performance across three different types of appliances including multi-state types which is harder to classify. 
    \item IMG-NILM proved robust when the number of appliances is increased.
\end{itemize}

This paper is structured as follows; a literature review of key techniques addressing NILM, with a specific focus on DNN approaches is discussed in Section \ref{LitRev}. The proposed image representation and  IMG-NILM architecture are presented in Section \ref{methodology}. Section \ref{ResAnal} reports the experiments and analysis of IMG-NILM. The paper is concluded in Section \ref{conclusion}.

\section{Literature Review}
\label{LitRev}
The research  on NILM was first popularised by George Hart in 1992 \cite{Hart} which focused mainly on the transitions between steady-state appliances. Traditional machine learning approaches have been explored for NILM. Probabilistic models such as in \cite{HMM} explored the possibilities of creating a hidden Markov model that can predict appliance use given active and reactive power as inputs. This technique, however, struggled to correctly identify appliances with multiple states, returning a high number of false negatives and false positives. An extension of this work is the popular methods of NILM disaggregation which uses factorial hidden Markov models (FHMMs) \cite{FHMM}. An FHMM can take into account multiple appliances with multiple running states. In an attempt to combat the challenges that arise when applying the FHMM to NILM, similar power signatures between appliances and the assumption of a uniform prior distribution are stated. Despite the improvement in accuracy, the model relies on a human-based analysis  that takes into account human-behavioural trends, which is not realistic on a larger scale.

Whilst these methods are often effective, the growing importance and complexity of the problems lend to a deep learning approach. With multiple variables, parameters and hidden patterns combined with new detailed, labelled datasets the training of a deep neural network has recently become a viable answer \cite{Kelly}. The time dependency and visual interpretations of the data can be explored in great detail using this approach. 

The majority of modern approaches to NILM disaggregation utilise deep neural networks. For instance,  
\cite{Kelly} depicted a sequence-to-sequence (seq2seq) method that adapted three different DNNs to fit a NILM model in the form of a `long short-term memory' (LSTM) architecture, a `denoising autoencoder'(DAE) and a CNN that regresses the start time, end time and average power demand for each appliance activation. LSTM networks are a form of recurrent neural network (RNN) meaning that there are feedback connections within their architecture which works as an `artificial memory'. The main function of an autoencoder is forced to reconstruct and reduce noise in the original data, a problem that aggregated data can cause, as is concurred by Siddiqui and Sibal \cite{Siddiqui}. LSTM is then applied to  classify the data based on trained data of power signatures. LSTM Showed a very good performance with appliances of two states but struggled to classify appliances which had multiple running states. Also, the method is computationally inefficient. 


 \cite{Pbnilm} used RNNs  with a combination of CNNs guided by a pinball loss function. This network is benchmarked not against similar architectures such as Seq2Point \cite{Seq2Point} and also against the same network with a different loss function (MSE). A pinball loss function is implemented to take into account appliances with unpredictable power patterns and this translated into very high accuracy levels when compared to other architectures as well as other loss functions. The major drawback of this work arises when attempting to disaggregate energy data coming from two appliances with similar energy outputs.
 
 Representing time series in NLIM as an image is a reasonably new concept and therefore a few papers in the literature have  specifically discussed it. \cite{kyrkou2019imaging} transformed energy consumption time series signal into polar coordinates and then utilised the Gramian Angular Summation/Difference Field (GASF/GADF) transformations to create 2-D images. The focus of this study is transfer learning when a  pre-trained VGG16 model is used for classifying appliances. This study showed that models using image representation have the potential to generalise better to new data, even when the data is from an entirely different dataset. \cite{senarathna2021image} compared various image representations for NILM, namely Gramian Angular Fields (GAF), Gramian Angular Difference Field (GADF) and Recurrence Plots (RP). The results showed that Gramian Angular Difference Field (GADF) outperforms other image representations. Although these representations demonstrated good potential for image representation in NILM, they are based on correlation rather than generated directly from the energy consumption, whereas the temporal aspect is preserved.  Also, most of the aforementioned methods focus on disaggregation of one type of appliance and do not perform well across different types. Another research that uses the heatmap representation but for genetic data is discussed in \cite{HeatmapPaper}.
 


\section{Methodology}
\label{methodology}

\begin{figure}[ht!]
    \centering
    \resizebox{0.5\textwidth}{!}{
    \includegraphics{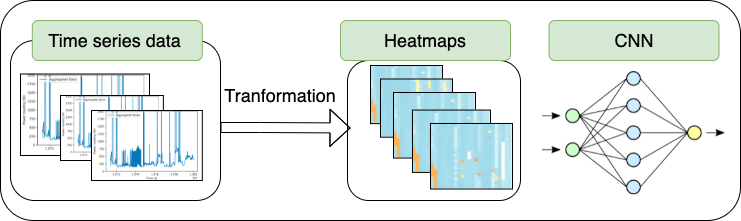}}
    \caption{Workflow of IMG-NILM }
    \label{overview}
\end{figure}

 IMG-NILM workflow starts with transforming the time series data into heatmaps which reflect the consumption of power over a specific window of time. This representation is key in the proposed method as it transforms the data representation from time series to an image, hence, this allows us to utilise state-of-the-art techniques in image classification, i.e., CNN, for NILM. An overview of the process of IMG-NILM is described in Figure \ref{overview}.
We first describe the proposed image representation of energy time series data which is the input for our model. Then, we illustrate the deep neural network architecture and parameters. 

\subsection{Image Representation}
\label{KnowRep}
\begin{figure}[ht!]
     \centering
     \begin{subfigure}{0.45\textwidth}
         \centering
         \includegraphics[width=\textwidth]{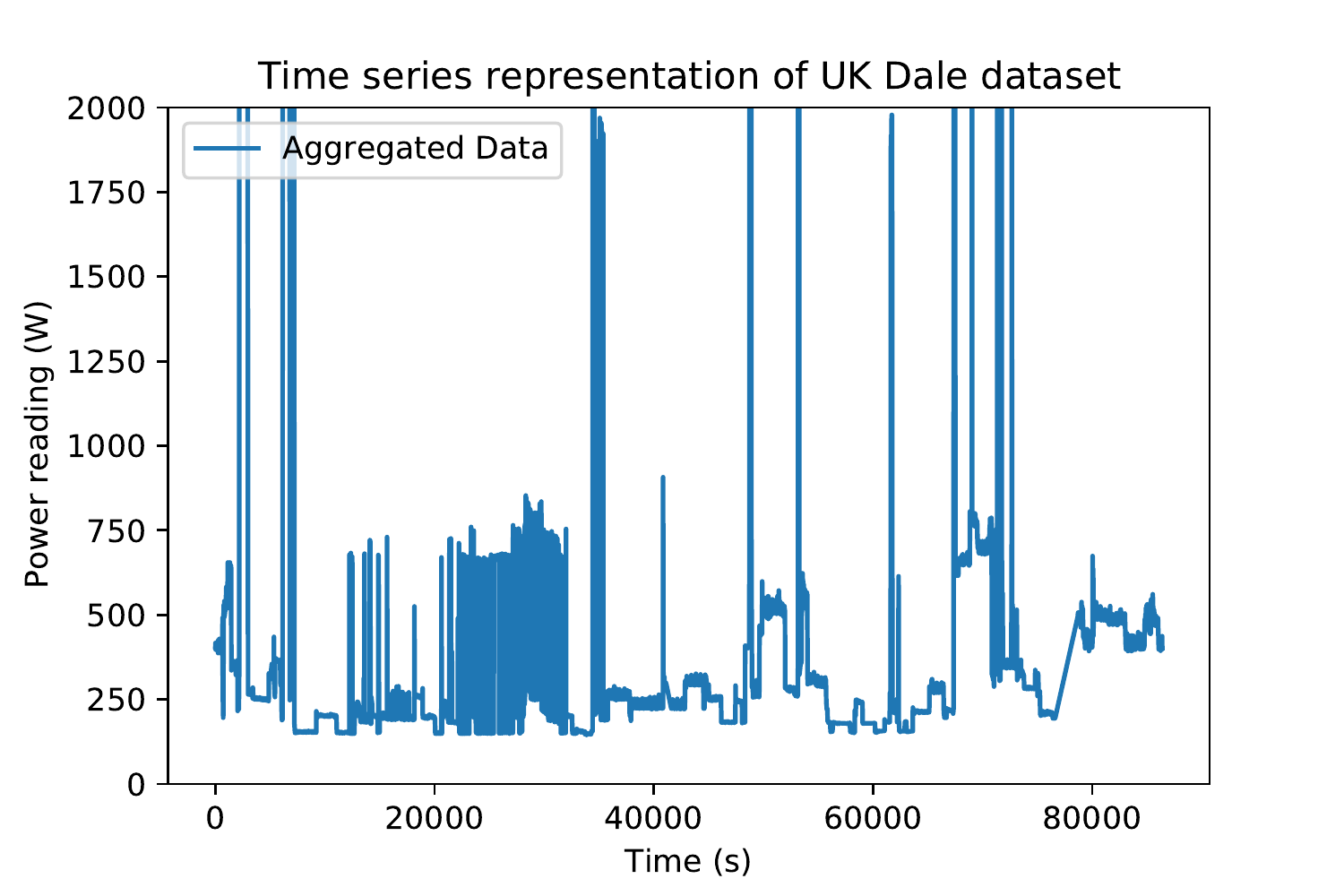}
         \caption{A snapshot of aggregated power data from a single house in the UK-Dale dataset}
         \label{TS_data}
     \end{subfigure}
     \hfill
     \begin{subfigure}{0.45\textwidth}
         \centering
         \includegraphics[width=\textwidth]{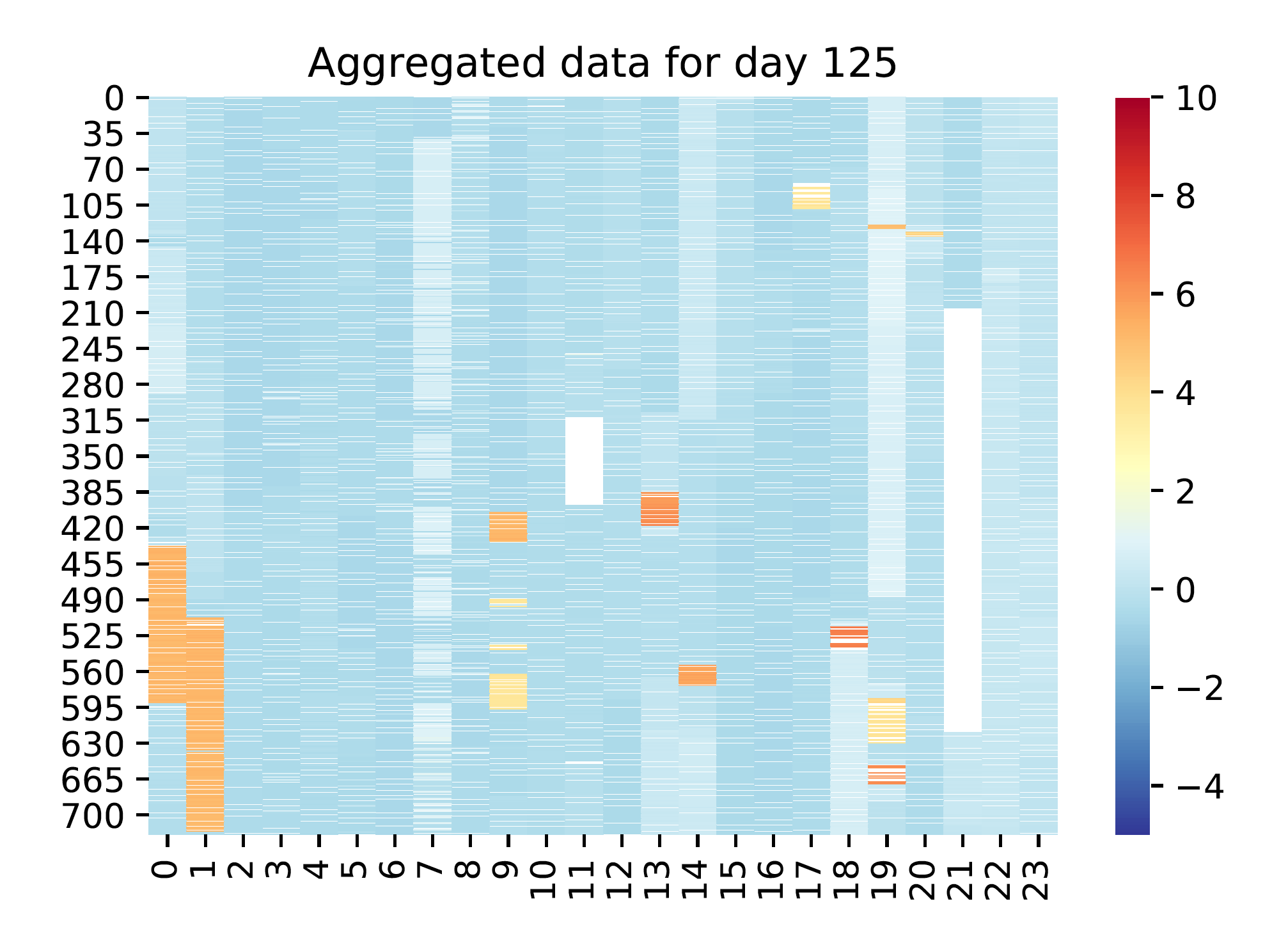}
         \caption{Heatmap depicting the total power readings for house 1 in the UK-Dale dataset (Day 125)}
         \label{HMAgg}
     \end{subfigure}
     \hfill
        \caption{The power consumption from a single house from the UK-Dale dataset in (a) time series format (b) heatmap format}
        \label{representation}
\end{figure}

Transforming data into a heatmap is a crucial step in IMG-NILM. In this step, an input of time series such as in Figure \ref{TS_data} is provided for a window of time $T$ to generate a heatmap such as in Figure \ref{HMAgg}. Each pixel in the heatmap corresponds to $n$ unit of aggregated data.

For a window of size $w$, the data is aggregated for each $i$ $\in$ $w$ according to an aggregation step $s$. For instance, in Figure \ref{HMAgg}, data is presented for a duration of a day, hence the window size;  $w$; is 24 (in hours unit). The aggregation step for this heatmap is specified as 5 seconds; $s=5$. Hence, for each $i$ in $w$ corresponds to an hour of the day, each pixel represents an aggregated data of $s$; 5 seconds, with a total of 720 pixels per hour (column). The data is normalised using z-score normalisation to standardise across different windows and houses. The heatmap depicts the consumption of power on a specific day with  dark blue colour showing low consumption that transitions through a light blue and into a red colour as the power values increase. The image shows clearly an increase in power consumption during evenings which could  be explained by turned-on heaters, with some scattered patterns relevant to routine-related activities such as cooking or watching TV mostly in the afternoon. 


 The goal of disaggregation using IMG-NILM is to differentiate between time periods when appliances were turned on/off. Hence, the classifier can decide whether a certain appliance is switched off from the heatmap of a specific window. This binary classifier will therefore require two classes of input images, one with the full aggregated data and the other with the total power reading without a chosen appliance's reading. By taking the power readings of these appliances from the overall aggregated power, a heatmap can be generated that exhibits the situation where the specific appliance is not present in the particular house or not on during a particular time window.


The input for the CNN is a collection of heatmaps representing the existence and absence of the target appliances. In the next subsection, we will discuss the details of the network deployed for this task. 

\subsection{Convolutional Neural Network}
Convolutional neural networks (CNNs) are typically used in image recognition. The convolution layers iterate a specified kernel over an RGB image (Red, green, blue), and from the resulting matrix, an input can be generated usually by summation for an activation function. These convolution layers are customarily followed by a pooling layer which reduces the size of the outputted image; a necessary adaptation that in turn depletes the number of parameters needed in the model. Convolutional layers tend to record the exact location of a feature, and this can reduce the generality of a model if the feature map is sensitive to the location of a pattern in the image. The resulting architecture tends to be a combination of these layers which specify input shape and kernel sizes.



\begin{figure}[ht!]
    \centering
    \includegraphics[width=0.4\textwidth]{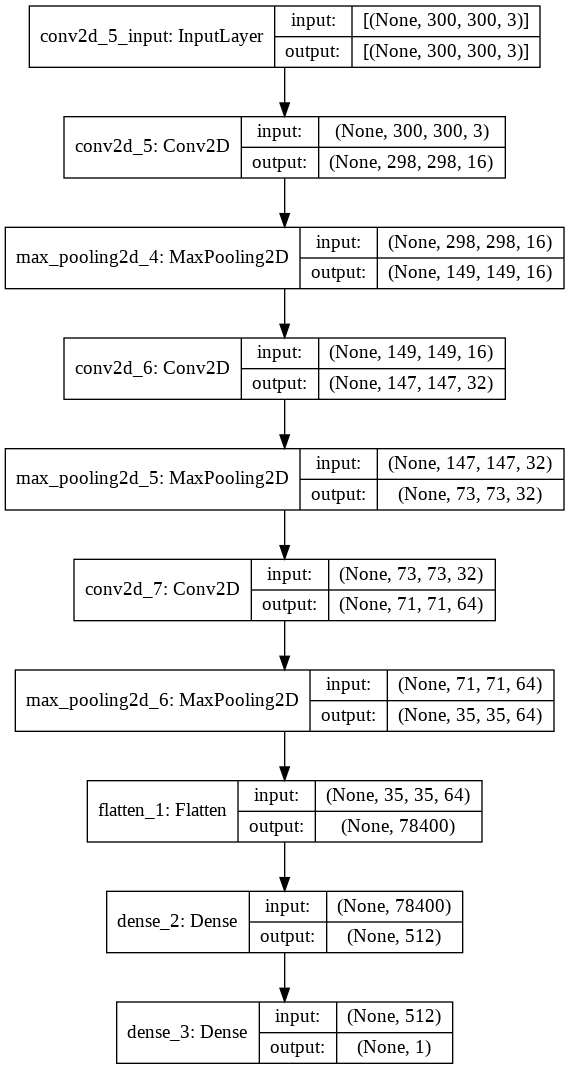}
    \caption{CNN architecture to predict whether a given appliance is on or off based on heatmap images}
    \label{fig:Devel1ARCH}
\end{figure}

The overall network architecture is described in Figure \ref{fig:Devel1ARCH}, and explained as follows:  
\begin{itemize}
    \item The input layer is in a form of a collection of heatmaps/images of size 300 $X$ 300 for three channels; representing RGB images. 
    \item The architecture is composed of three consecutive convolutional layers, each followed by a pooling layer to perform down-sampling of the spatial dimension of the input.
    \item The convolutional layers are followed by batch normalisation \cite{ioffe2015batch} to improve learning and Rectified Linear Unit (ReLU).
    \item The network is finalised with three fully connected layers, all of which utilise a \textit{softmax} activation function. Naturally, this model is paired with a categorical cross-entropy to detect the presence or absence of an appliance from an aggregated heatmap.
\end{itemize}

It is important that the model does not overfit certain images especially when training on data from different houses. Dropout layers are commonly used to counteract a model over-fitting the training data. By setting the inputs of a certain percentage of nodes, dropout layers ensure that the tuning of these parameters does not over-compensate across the whole network. Hence, when applying IMG-NILM on data across houses, a single dropout layer is added after the final pooling layer with a dropout rate of 0.25. 

\section{Experimental Analysis}
\label{ResAnal}

Household appliances are categorised into four types. Type I appliances are the most common, with two states, on and off. Type II appliances have multiple running states such as dishwashers. Type III have an infinite number of running states and type IV are those appliances that remain on throughout the day such as a refrigerator or freezer. Understanding these types allows us to effectively select appliances that we can focus on evaluating our proposed method. We focus in this paper on the three most common types of I, II and IV. A Type IV appliance will have a constant impact on the total power reading of a house, while type I appliances have a distinguished signature of step up and step down the aggregated power simultaneously with the on/off button being pressed. Type II has a similar behaviour as type I, yet with an extension for multiple states.

In this paper, we use the UK-Dale dataset to evaluate the performance of IMG-NILM and compare it to benchmark methods. The experiments assess the desegregation performance for example appliances in three  different types; Type I (on and off states), Type II (appliances with multiple on and off states) and Type IV (always on with different cycles). These experiments aim to answer the following questions: 
\textbf{Q1}: How the accuracy of IMG-NILM differs across Types? \textbf{Q2}: How the IMG-NILM performs with data from a single house with a large number of appliances operating simultaneously? \textbf{Q3}: What is the impact of using data across different houses for training and testing?
We first explain the dataset and the experiment setup. This will be followed by a discussion of the experiments addressing the aforementioned questions.  
\subsection{Data}
\label{data}

For the analysis we use the UK-Dale dataset \cite{UKDaleData}. The data has over 140,000,000 seconds worth of data (over 1,620 days) and 5 separate houses with measurements roughly every 6 seconds. The combination of high frequency and large size means that passing a moving daily window across different houses would yield enough images to input into our method. House 3 is excluded in this analysis as it lacks data for the three target types. House 1 is particularly interesting in this dataset as it includes 3.5-year power data for 52 appliances. Hence, disaggregation of a single appliance from aggregated data is expected to be challenging for this house. The number of appliances in other houses is smaller with 20,5,6 and 26 for houses 2,3,4 and 5 respectively. In the experiments when analysing data from the same house, we choose house 1 for this task as it contains the largest set of appliances, hence more challenging. This will be followed by using data across different houses for training and testing. 

\subsection{Experiments setup}
Two parameters are to be specified for creating the heatmaps as explained in Section \ref{methodology}. The first is the window size $w$; which reflects the window of time for the heatmap. $w$ is set in all experiments to one day (24 hours). The other parameter is the aggregation step $s$, which is set to be 6 seconds. An RGB heatmap of size 720x24x3 is generated for each day. The total number of images generated for experiments is stated in Table \ref{images}.


\begin{table}[ht!]
\centering
\caption{Total number of images generated from the UK-Dale dataset across three types of appliances Type I (TV), Type II (dishwasher) and Type IV (refrigerator) }
\begin{tabular}{|l|l|l|l|}
\hline
Dataset   & Dishwasher & TV & Refrigerator \\\hline
House 1 & 1995    & 1400     & 2565  \\ \hline
Across houses     & 2418     &    1800   &3080    \\ \hline
\end{tabular}

\label{images}
\end{table}

\begin{figure}[ht!]
     \centering
     \begin{subfigure}{0.45\textwidth}
         \centering
         \includegraphics[width=\textwidth]{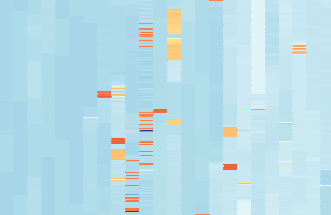}
         \caption{Class I}
         \label{fig:classI}
     \end{subfigure}
     \hfill
     \begin{subfigure}{0.45\textwidth}
         \centering
         \includegraphics[width=\textwidth]{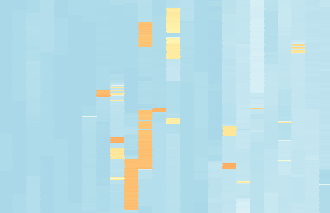}
         \caption{Class II}
         \label{fig:classII}
     \end{subfigure}
     \hfill
        \caption{An example of the generated heatmaps where (a) Class I is where the dishwasher power signature is absent from the aggregated data (b) Class II: Aggregated data}
        \label{ClassExample}
\end{figure}

Various heatmaps are generated to classify three types of appliances across three types; Type I (TV is an example), Type II (dishwasher is an example) and Type IV (where the refrigerator is an example). The generated images are split into two classes; \textit{Class I:} contains heatmaps which have either been produced from days where the appliance was not switched on at all, or are generated from removing that appliance's power signature from the total aggregated signature, thus summing to a class which does not contain the selected pattern's `heat signature'.\textit{Class II:} on the other hand, solely accommodates heatmaps that represent days when the chosen appliance has been on, therefore showing the `heat signature' that reflects the selected appliance. Examples of images in the two classes at the same time window are presented in Figure \ref{ClassExample}

When assessing neural networks there is a multitude of performance metrics that one can employ to highlight the accuracy and robustness of the particular model. A commonly used metric is the training/validation/test accuracy of the model. Accuracy is calculated by:

\begin{equation}
    Accuracy = \frac{TP + TN}{TP + TN + FP + FN}
\end{equation}

Accuracy determines the percentage of correct predictions for the test data. Here, TP is true positive, FP is false positives, TN is true negatives and FN is false negatives. True positives and true negatives are the observations that are correctly predicted while false positives and false negatives are observations where the actual class differs from the predicted class.

We run our experiments with an 80:20 ratio for the training:test split. With a 20\% split for the validation set of the training set. The heatmaps are automatically shuffled within their respective classes so that the training/validation images are selected from a random sample of days. All of the models make use of the `adam' optimiser with a loss rate of $1\times10^{-4}$.

\subsection{Experiments on a single-house data}

\begin{figure}[ht!]
\centering
\centering
\includegraphics[width=0.5\textwidth]{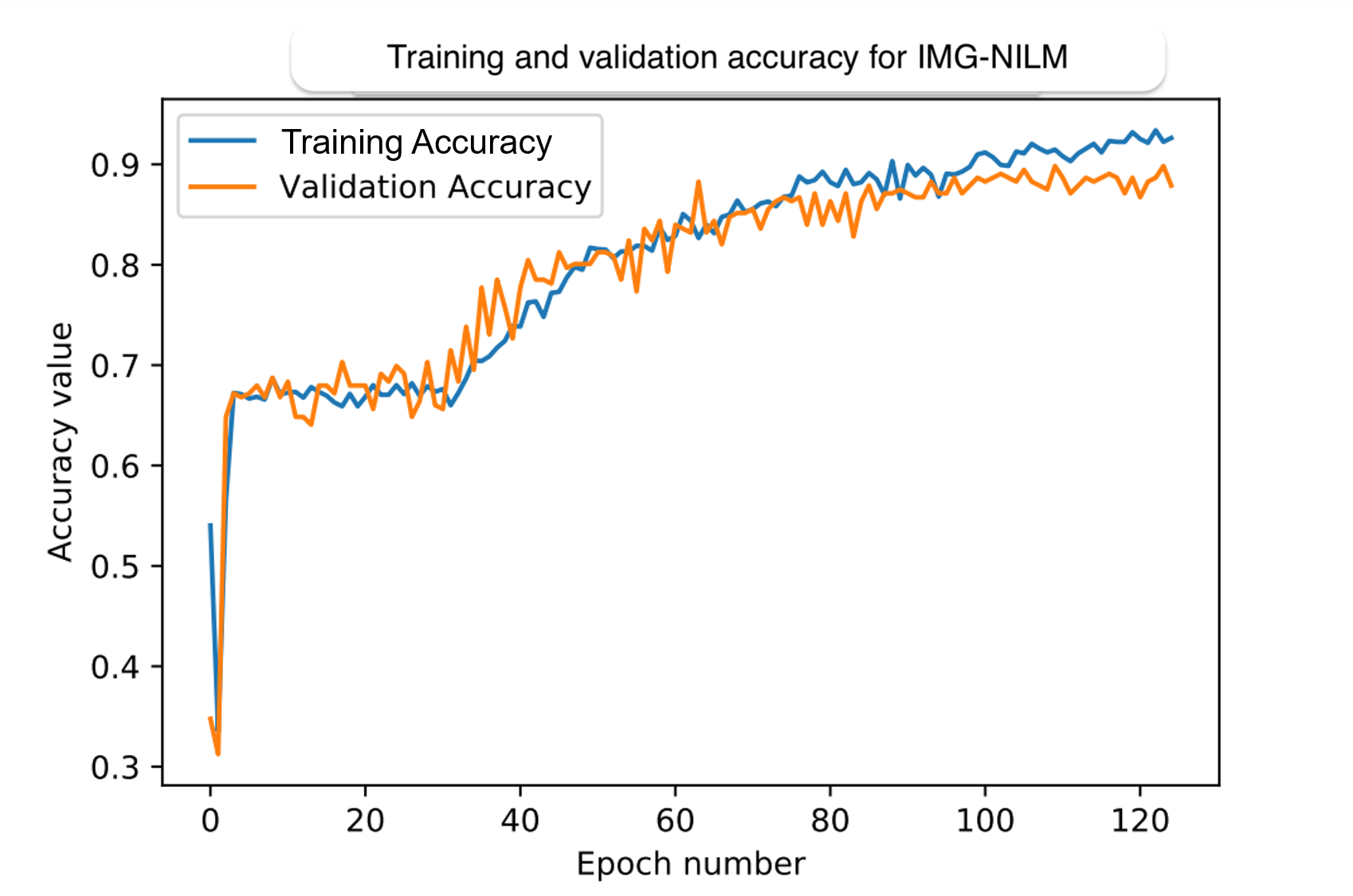}
\caption{Training and validation accuracy for Type II appliance (dishwasher) with respect to the epochs on data from a single house}
\label{epochs_house1}
\end{figure}

We address in this section Q1: How does the accuracy of IMG-NILMdiffers across Types? and Q2: How does IMG-NILM perform with data from a single house with a large number of appliances operating simultaneously? We start with data from a single house, where the model is trained to learn certain patterns that the CNN identifies and therefore return whether a certain appliance is on or off. This particular house has a large number of appliances (512) across the four types, hence the classification is challenging. Figure \ref{epochs_house1} shows the training and validation accuracy with different epochs. 

These graphs visualise the process that the model goes through when attempting to maximise the training/validation accuracy. In the instance of the dishwasher classes, both the training accuracy and the validation accuracy with scores of 92\% and 88\% respectively. The test results further back up the validity of this test with a test accuracy value of 85\%.  Dishwashers, a Type II appliance, have irregular but strong power signatures, using a lot of electricity for a specified length of time. Due to this, when converted into a heatmap image, the dishwasher provides a specific `heat signature' which displays itself as a pattern on the image. 

However, refrigerators (Type IV) are, in general, always turned on. Their power output remains somewhat constant and therefore, when this appliance is turned off, the impact on the heatmap is mainly on reducing the colour intensity of every pixel across the whole day. A CNN looking for patterns within the images will therefore struggle to find any similar `heat signatures' that would identify the presence/absence of a refrigerator. The challenging task for the classifier is, in some respects, searching for different keys as to how to differentiate between the two classes in each experiment; pattern identification in the dishwasher set and a colour difference in the refrigerator set.

\begin{figure}[ht!]

\centering
\includegraphics[width=0.5\textwidth]{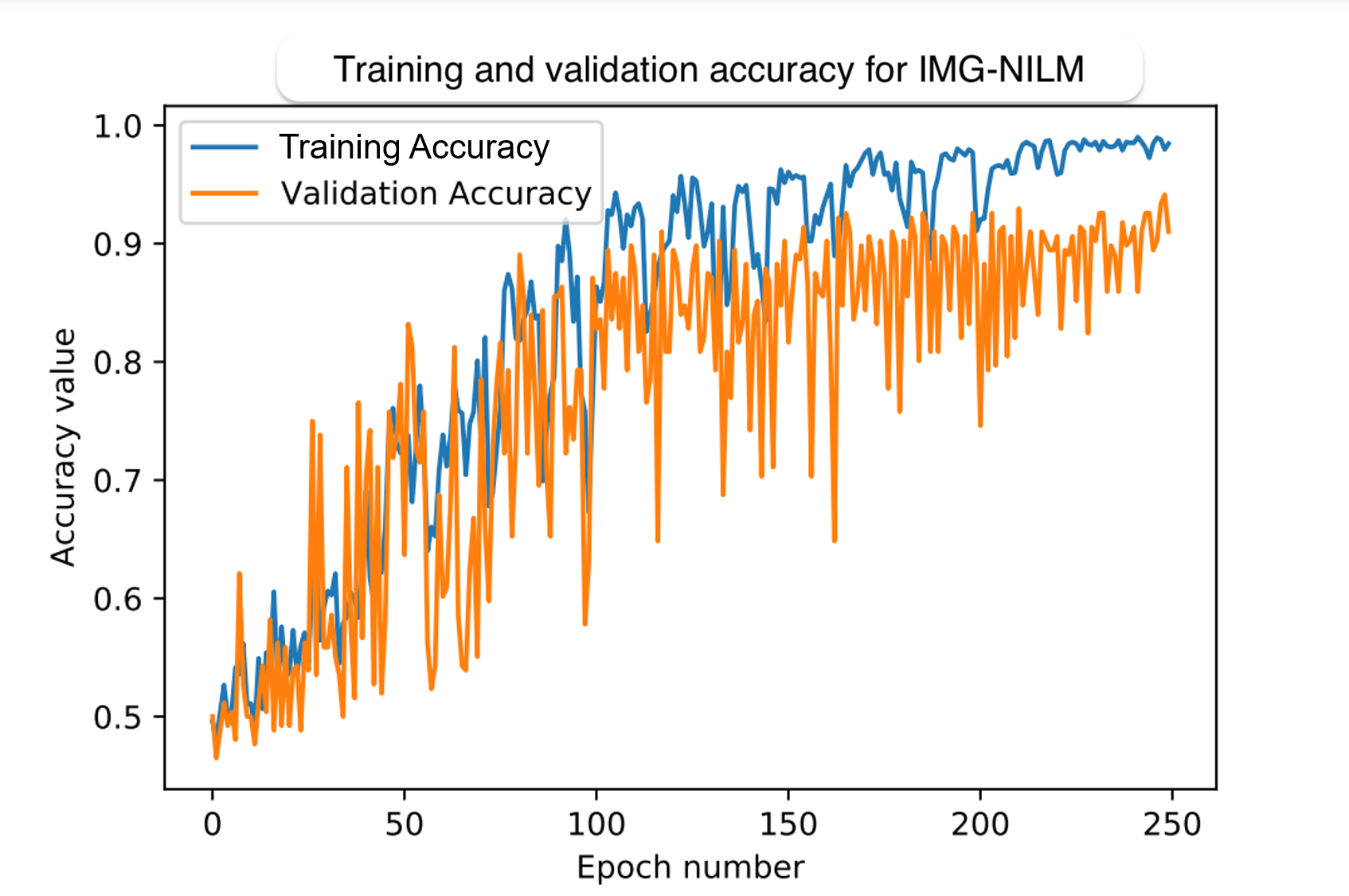}
\caption{Training and validation accuracy for Type IV appliance(refrigerator) with respect to the epochs on data from a single house}

    \label{fig:Fr_devel2}
\end{figure}

The number of epochs is adjusted to capture the difference between appliances, hence enhancing the classifier accuracy. Increasing the epochs can possibly lead to over-fitting, yet, in this model, when the epoch value is increased to 250, as shown in Figure \ref{fig:Fr_devel2}. The figure also demonstrates  the learning  across 125 epochs struggles to consistently classify the heatmaps that relate to the refrigerator with an accuracy level of over 80\% and when tested on unseen images, returns a test accuracy of just 71\%, which is significantly enhanced with more epochs. 

\begin{figure}[ht!]
    \centering
    \includegraphics[width=0.5\textwidth]{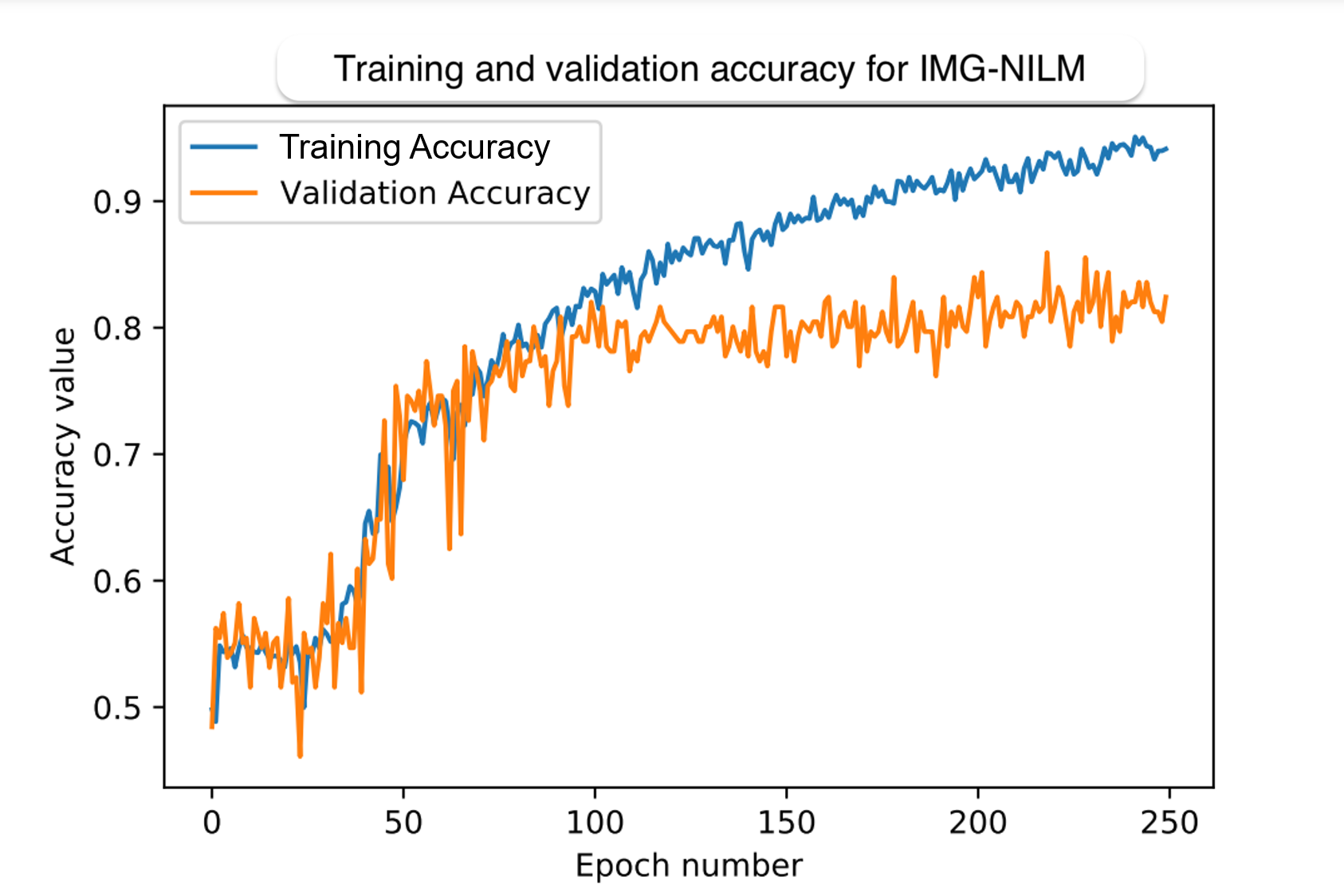}
    \caption{Training and validation accuracy for Type I appliance(TV) with respect to the epochs on data from a single house}
    \label{TV_house1}
\end{figure}

Figure \ref{TV_house1} shows the training and validation accuracy for TV which represents another type of appliance, Type 1 with only two states. The training phase of this model (250 epochs) shows a great jump in accuracy at around 50 epochs and from there slowly refined the accuracy values until results of 94\% and 82\% were finally output for training accuracy and validation accuracy respectively. When tested on unseen images in a test set of 411 images, the CNN attained an accuracy of 86\%. 

A summary of the results across the three appliances in House 1 data is shown in Table \ref{Model_Summary}. Training accuracy is very high across all appliances, however in testing, classifying refrigerators achieves the best accuracy, while TV and Dishwasher are less accurate at 86\% and 85\% respectively. 

\begin{table}[ht!]
\centering
\caption{Summary of accuracy on UK-Dale house 1 dataset across three Types of appliances}
\begin{tabular}{|l|l|l|l|}
\hline
Appliances   & Training & Validation & Testing \\\hline
Refrigerator (Type IV)& 98\%     & 91\%       & 93\%  \\ \hline
TV (Type I)          & 94\%     & 82\%       & 86\%    \\ \hline
Dishwasher (Type II) & 92\%     & 88\%       & 85\%    \\ \hline
\end{tabular}

\label{Model_Summary}

\end{table}
These results show the robustness of the proposed model across different appliances demonstrating the model's ability to learn different patterns of diverse types. Despite a large number of appliances in the aggregated images in this house (52 appliances), IMG-NILM can detect devices of different types with high accuracy. A multi-state device (dishwasher) is showing a robust performance, very close to Type I, which is a non-trivial task in NILM. 

\subsection{Experiments on across-houses data}

\begin{figure}[!ht]
    \centering
    \resizebox{0.5\textwidth}{!}{
    \includegraphics{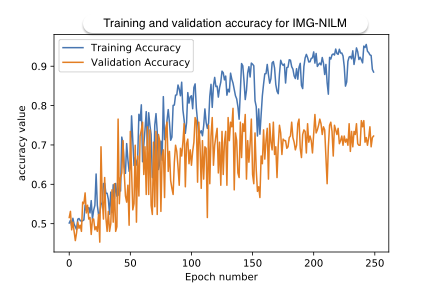}}
    \caption{Training and validation accuracy of a dishwasher classifier with respect to the epoch number on across-houses dataset}
    \label{fig:DW_devel3}
\end{figure}

\begin{figure}[h!]
    \centering
    \resizebox{0.5\textwidth}{!}{
    \includegraphics{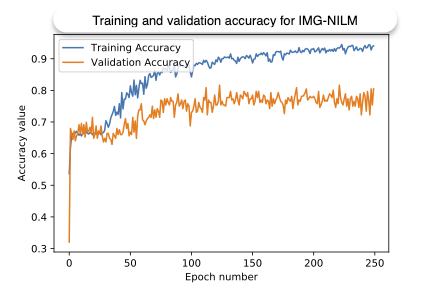}}
    \caption{Training and validation accuracy of classes relating to refrigerator data with respect to the epoch number on across-houses dataset}
    \label{fig:Fr_devel3}
\end{figure}

In the following experiments, we look to expand on our data to include heatmaps from different houses (addressing Q1 and Q3). 
Adding the data from the additional houses results in a harder task with complicated heatmaps. To avoid over-fitting, A drop-out layer is added with a drop-out rate specified as 0.25 in the following experiments. In Figure~\ref{fig:DW_devel3} we can see the training and validation accuracy using data from across houses as the number of epochs rises. The 0.25 dropout rate indicates that whilst the parameters and hyper-parameters continue to be calculated and recalculated, the random dropout of the nodes means that they cannot over-fit. The accuracy of training and validation for dishwashers are shown in Figure~\ref{fig:Fr_devel3}. The proposed method finishes training over 250 epochs with a training accuracy of 94\%, a validation accuracy of 80\% and a testing accuracy equalling 87\%. 

When comparing the results of the refrigerator, in Figure \ref{fig:Fr_devel3}, to the dishwasher classifiers, in Figure \ref{fig:DW_devel3}, it is evident that type IV appliances are more challenging to classify. The validation accuracy remains low with an eventual value of 74\%. When shown previously unseen test images, the network correctly classifies 84\% of them. 

Table \ref{Model_Summary2} summarises training, validation and testing accuracy for each appliance representing the three types. As excepted when working with data from across houses, validation accuracy has dropped from what has been reported in Table \ref{Model_Summary} when using data from a single house. However, IMG-NILM still attains a consistently high accuracy across different types of appliances. 

It is paramount therefore that the model does not overfit a particular house, especially since no two homes are the same in energy usage. One of the variables that has a major impact on the power output of a home is the brands and models of everyday appliances, with different models of the same appliance often showing different heat signatures altogether. In this case, the first house in the UK-Dale dataset uses a \textit{Whirlpool DWH B10} bought in 2007 whereas both house 2 and house 5 are fitted with a \textit{EcoManagerTxPlug}~\cite{UKDaleData}. These small differences can lead to slight inconsistencies in the input heatmap images. Nevertheless, IMG-NILM shows robustness when tested across houses. 

\begin{table}[ht!]
\centering
\caption{Summary of the accuracy of the UK-Dale dataset on different houses across three Types of appliances}
\begin{tabular}{|l|l|l|l|}
\hline
Appliance   & Training & Validation & Testing \\ \hline
Refrigerator (Type IV)& 89\%     & 74\%       & 84\%    \\ \hline
TV (Type I)      & 93\%     & 69\%       & 83\%    \\ \hline
Dishwasher (Type II)  & 94\%     & 80\%       & 87\%    \\ \hline
\end{tabular}

\label{Model_Summary2}
\end{table}
Comparing our method to the literature has encountered multiple challenges. For example, \cite{Pbnilm} used a slightly different metric, known as \textit{correctly estimated power (CEP)}, the three models presented in the paper return modal CEP values of 94\%, 97\% and 66\% when looking at a fridge-freezer and 89\%, 98\% and 67\% for a dishwasher. The paper itself declares however that this does not reflect the overall accuracy of the model as the test metric only reports the performance concerning the rate of success. Also, they use other datasets to test the performance. If we take their results as guidance when comparing IMG-NILM performance, we see very similar values. \cite{DLandPP} reported estimated test accuracy of 89\% when attempting to predict the presence of a dishwasher. However, the design of the models in the discussed paper is specifically structured to work effectively on Type II appliances. Although their results are slightly higher than IMG-NILM (87\% for dishwashers), IMG-NILM is not exclusive to one type and can be applied to various types of appliances, while still maintaining good accuracy.   

A relevant image-based NILM method is presented in \cite{senarathna2021image}. However, this has been only tested for one house on the REDD dataset by \cite{redd}, which is much smaller and has a fewer number of appliances. Also, the study looked at refrigerators only. The reported validation accuracy is between 82.7 \% and 94.2\%. The comparable accuracy we report with IMG-NILM on a dishwasher tested on a single house is 91\% validation accuracy and 93\% for testing. It is also noted that the data from this house is longer and more complex due to the large number of appliances that exist.

\section{Conclusion and future work}
\label{conclusion}
 In this paper, we introduce IMG-NILM which adapts the CNN architecture for energy disaggregation based on images, i.e., heatmaps, generated from power consumption. The proposed approach shows an accurate and robust performance in challenging scenarios, both in a single house when many devices are operating and also across houses. IMG-NILM can be successfully used for different types of appliances with single or multiple states, constant or variant power consumption. 

Future work will focus on extending the model to detect multiple appliances at the same time, which is one of the main challenges in NILM. Also, we will focus on investigating the parameters for generating images and the best window for each appliance.

\bibliographystyle{unsrtnat}
\bibliography{references}  






\end{document}